\documentclass[letterpaper, 10 pt, conference]{ieeeconf}  %

\IEEEoverridecommandlockouts                              %

\usepackage{amsmath}

\usepackage{accents}

\makeatletter
\def\BState{\State\hskip-\ALG@thistlm}
\makeatother

\usepackage{amssymb}

\usepackage{algorithm}
\usepackage[noend]{algpseudocode}

\usepackage{mathrsfs}
\usepackage{mathtools}
\usepackage{amsfonts}

\usepackage{subcaption}

\usepackage{stackengine}
\setstackgap{S}{2pt}

\usepackage[
    colorlinks,
    urlcolor=black,
    citecolor=black,
    linkcolor=black,
    pdfpagemode=UseOutlines, %
  ]{hyperref}
  
\usepackage[capitalise]{cleveref}
\usepackage{xcolor}

\usepackage{siunitx}

\usepackage{multirow}
\usepackage{cite}
\usepackage{balance}
\usepackage{url}

\title{\LARGE \bf

	Solving Sequential Manipulation Puzzles by Finding Easier Subproblems
}

\author{Svetlana Levit$^{1, 2}$, Joaquim Ortiz-Haro$^{1,3}$ and Marc Toussaint$^{1, 2}$%
\thanks{This research has been supported by the German Research Foundation (DFG) under Germany's Excellence Strategy – EXC2002/1–390523135 "Science of Intelligence",
and the German-Israeli Foundation for
Scientific Research (GIF), grant I-1491-407.6/2019.}%
\thanks{Correspondence to levit@tu-berlin.de}
\thanks{$^{1}$Learning \& Intelligent Systems Lab, TU Berlin, Germany
        }%
\thanks{$^{2}$Science Of Intelligence Cluster of Excellence}%
\thanks{$^{3}$Machines in Motion Laboratory, New York University, USA}%
}

\begin{document}

\maketitle
\thispagestyle{empty}
\pagestyle{empty}

\begin{abstract}
We consider a set of challenging sequential manipulation puzzles, where an agent has to interact with multiple movable objects and navigate narrow passages. Such settings are notoriously difficult for Task-and-Motion Planners, as they require interdependent regrasps and solving hard motion planning problems.

In this paper, we propose to search over sequences of easier pick-and-place subproblems, which can lead to the solution of the manipulation puzzle.
Our method combines a heuristic-driven forward search of subproblems with an optimization-based Task-and-Motion Planning solver. 
To guide the search, we introduce heuristics to generate and prioritize useful subgoals.
We evaluate our approach on various manually designed and automatically generated scenes, demonstrating the benefits of auxiliary subproblems in sequential manipulation planning.
\end{abstract}

\section{Introduction}

Manipulation planning in the presence of movable objects and an obstructed environment is challenging: the number of possible action sequences grows with the number of objects, leading to a hard combinatorial problem, whereas obstacles and narrow passages make motion planning difficult. 
Sequential manipulation requires reasoning about what objects to interact with, how to grasp, and where to place them.
At the same time, collision constraints with the environment may introduce interdependencies between grasps, placements of objects, and robot trajectories, making planning especially hard.

\begin{figure}[t]
\centering
    \begin{subfigure}[t]{.21\linewidth}
        \includegraphics[width=.97\linewidth]{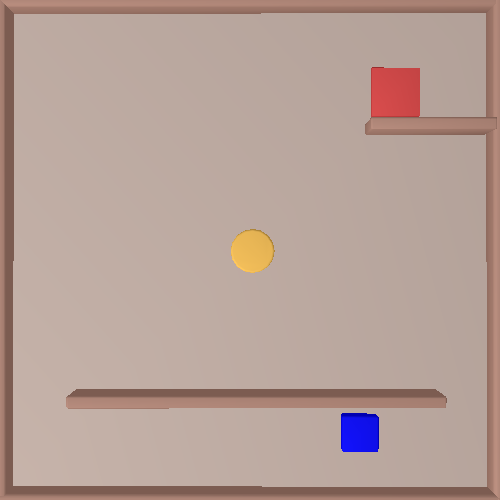}
        \caption{}
        \label{fig:p3a}
    \end{subfigure}
    \begin{subfigure}[t]{.21\linewidth}
        \includegraphics[width=.97\linewidth]{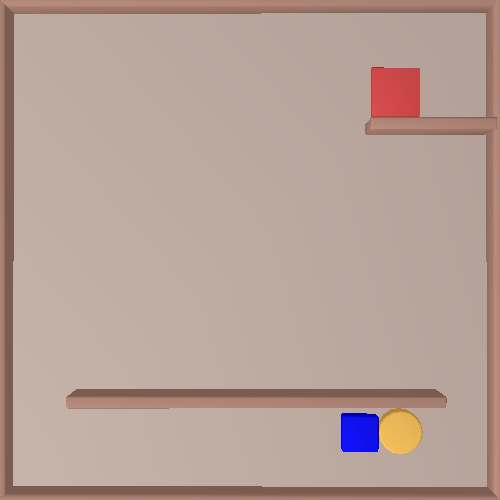}
        \caption{}
        \label{fig:p3b}
    \end{subfigure}
    \begin{subfigure}[t]{.21\linewidth}
        \includegraphics[width=.97\linewidth]{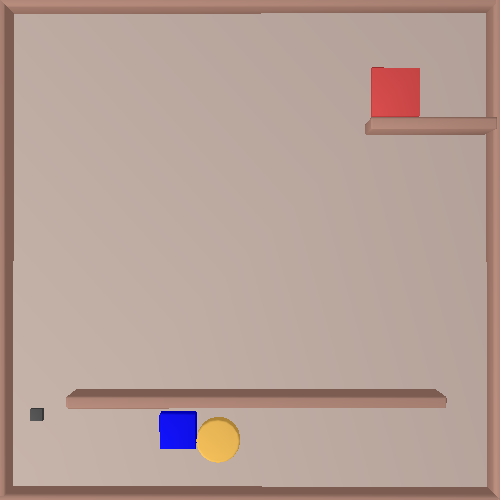}
        \caption{}
        \label{fig:p3c}
    \end{subfigure}
    \begin{subfigure}[t]{.21\linewidth}
        \includegraphics[width=.97\linewidth]{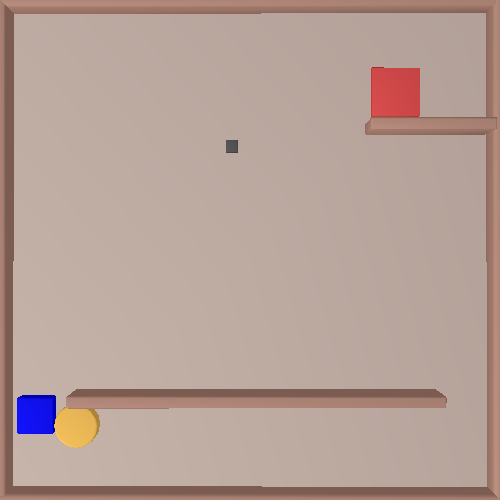}
	\caption{}
        \label{fig:p3d}
    \end{subfigure}
    \begin{subfigure}[b]{.21\linewidth}
        \includegraphics[width=.97\linewidth]{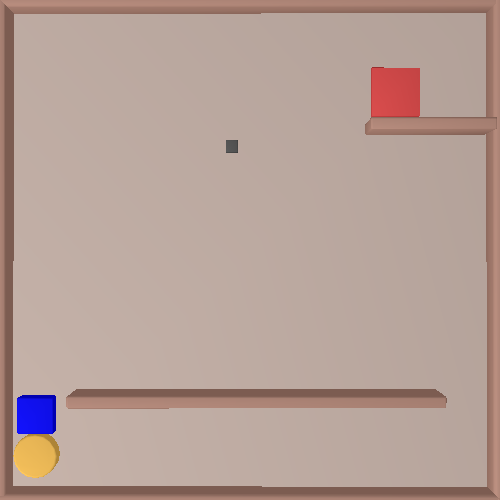}
        \caption{}
        \label{fig:p3e}
    \end{subfigure}
    \begin{subfigure}[b]{.21\linewidth}
        \includegraphics[width=.97\linewidth]{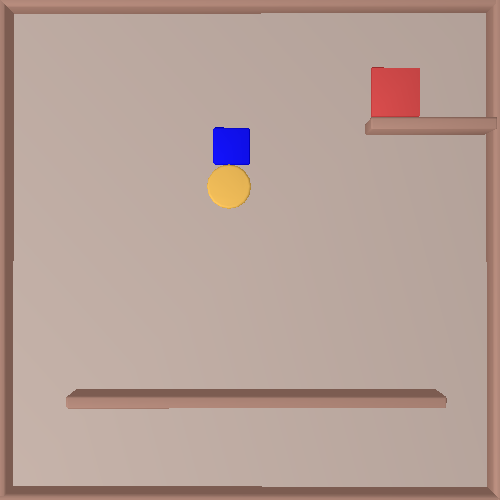}
        \caption{}
        \label{fig:p3f}
        \end{subfigure}
    \begin{subfigure}[b]{.21\linewidth}
        \includegraphics[width=.97\linewidth]{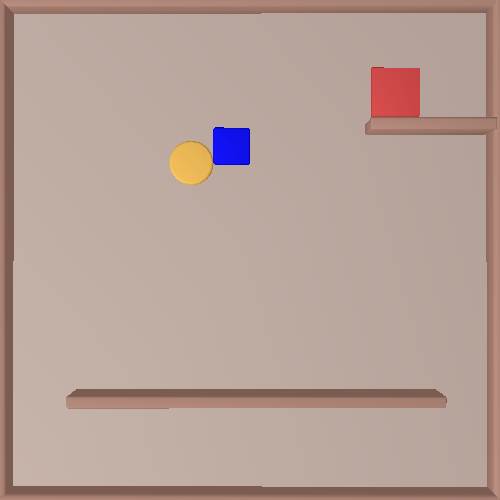}
        \caption{}
        \label{fig:p3g}
    \end{subfigure}
     \begin{subfigure}[b]{.21\linewidth}
        \includegraphics[width=.97\linewidth]{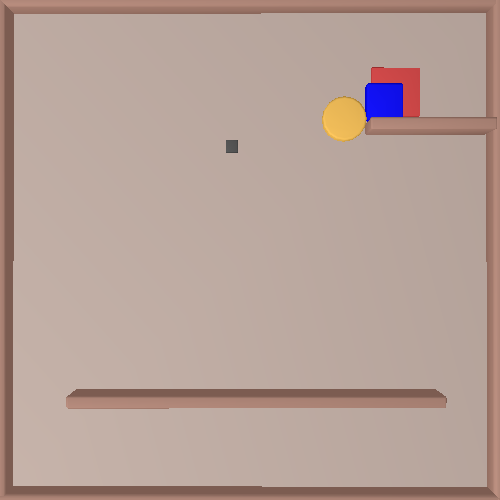}
        \caption{}
        \label{fig:p3h}
    \end{subfigure}
	\caption{Solving the Wall pick-and-place puzzle with subgoals. In order to bring the goal-object (blue) to the goal (red square), the agent (yellow circle) moves it and regrasps at several intermediate subgoal locations (gray dots).}
    \label{fig:puzzle3solution}
\end{figure}

Consider the puzzle in Fig.\,\ref{fig:puzzle3solution}, where the 2-DoF robot agent (the yellow circle) has to move the blue cube into the red goal area. Due to the narrow entrances the cube can only be retrieved from behind the wall if the agent grasps it from the bottom or the top, which is not possible at the current position. Multiple regrasps are necessary to solve this problem: e.g.\ move the cube inside the left entrance (\ref{fig:p3b}-\ref{fig:p3d}), regrasp from the bottom (\ref{fig:p3e}), move to the center (\ref{fig:p3f}), regrasp and move to the goal (\ref{fig:p3g}-\ref{fig:p3h}).

As shown on this example, solving such sequential manipulation problems requires regrasping an object at different angles and positions, finding good intermediate placements for the objects, and planning trajectories.
Such problems can be formulated as Task and Motion Planning. However, they pose the following challenges to TAMP planners:
\begin{itemize}
    \item Motion planning with spatial non-convexities is a challenging problem by itself.
    \item For optimization-based TAMP approaches the non-convexity could lead to local optima.
    \item Manipulation planning is especially hard for narrow passages, where only very few grasp positions and orientations lead to feasible downstream paths and manipulation.
\end{itemize}
To address these challenges, we propose to solve the full TAMP problem by searching sequences of auxiliary pick-and-place subproblems, where objects need to be placed at intermediate locations: the subgoal locations.
We introduce heuristics to guide the subproblem search and present several methods to generate subgoals for the pick-and-place subproblems. We evaluate their performance and capability to handle scenes with obstacles and narrow passages.

To summarize the contributions of our work, in this paper we present:
\begin{itemize}
\item A forward search method for solving sequential manipulation problems, integrating a subproblem search with an optimization-based TAMP solver.
\item Two approaches to generate and select subgoal locations that serve as regrasp location candidates in a sequential manipulation problem.
\item Heuristics to find configurations feasible for the optimization-based TAMP solver and useful to reach the goal.
\item A set of pick-and-place puzzles which we created to evaluate our solver, together with solutions and performance metrics.
\end{itemize}
\section{Related Works}

Many challenges of our problem setting are inherent to Task-and-Motion Planning (TAMP) problems due to the dependencies between discrete logical decisions and continuous physical states. Recent TAMP approaches are thoroughly reviewed in \cite{Garrett2020}.
TAMP solvers differ especially in the way physical constraints and motion planning are integrated into the task planning.
Sample-based TAMP planners incrementally discretize the continuous configuration space. For instance, PDDL-Stream \cite{garrett2020pddlstream}, combines sequential constrained sampling with PDDL Planning, whereas \cite{ferrer2017combined} 
integrates precomputed samples of feasible configurations into task planning. 

Using constraint propagation techniques, physical constraints can be incorporated in task planning to allow informed decisions \cite{lagriffoul2014efficiently,bidot2015geo,perez2014constraint}.
Another approach to solve TAMP problems, instead of sampling, is to use optimization methods to resolve dependencies and plan with nonlinear and geometric constraints \cite{Toussaint2015,ortiz22conflict,ortiz2022conflictinterface}.
Most TAMP solvers focus on long-term planning for manipulating multiple objects in a static environment with few environment constraints, such as placing objects on big tables, with informative goals or task descriptions.
In contrast, our problems require shorter action sequences with very precise manipulation, removing potential obstacles, and navigating narrow passages, without any guidance from the symbolic goal.
We extend the optimization-based TAMP solver in \cite{Toussaint2015} with a forward search over smaller subproblems, which can be solved with a sampling based motion planner.
Closely related to TAMP, multi-modal planning defines manipulation planning as a hybrid planning with grasps seen as modes\,\cite{Hauser2011,Vega2020,Kingston2020a,schmitt2017optimal}.
To handle manipulation in constrained spaces, \cite{Simeon2004} explores the idea of probabilistic roadmaps for planning of grasps and placements. However, they only consider a single movable objects and do not include symbolic task planning. Similarly related is rearrangement planning, addressed by \cite{Ota2004,wang2022efficient,ren2023kinodynamic,bayraktar2023solving}, with the important distinction that we do not know the target positions of any additional objects.%

The problem settings closest to ours are Manipulation Among Movable Obstacles\cite{Saxena2021ManipulationPA,stilman2007manipulation} and Navigation Among Movable Obstacles\cite{Stilman2008}, which can also be considered a special case of TAMP. 
These approaches often assume \textit{monotone} manipulation cases, where every object is grasped only once. Many cluttered settings, including some of our puzzles, require multi-step interactions and finding intermediate positions for objects, before moving them to the goal. %

The concept of subgoals has been extensively discussed in Reinforcement Learning (RL), e.g.\ \cite{chane2021goal,nasiriany2019planning,Fei2021,jurgenson2020a}.
While these works do not address TAMP, they provide interesting ideas for proposing subgoals.
In particular, diverse density analysis \cite{McGovern2001AutomaticDO} is an approach to characterize bottleneck regions, on which our method builds to propose subproblems.

In classical planning, landmarks 
\cite{hoffmann2004ordered}, 
(related to our definition of subgoals) are used to guide search-based planning \cite{PEREIRA2020, Karpas2009CostOptimalPW},
but are defined in the discrete logic domain.
We draw on such ideas and transfer them to proposing manipulation subproblems for TAMP solving.
\section{Problem Formulation and Notation}
\newcommand{\RRR}{{\mathbb{R}}}
\newcommand{\XX}{{\mathcal{X}}}
\newcommand{\1}{{\text{-1}}}
\subsection{Logic Geometric Programming for TAMP}
We adopt an optimization-based formulation of Task-and-Motion Planning, namely as Logic-Geometric Program (LGP)\cite{Toussaint2015}\cite{Toussaint2018}.
Given the configuration space $\XX=\RRR^n \times SE(3)^m$ of a $n$-DoF robot and $m$ movable objects, LGP assumes that feasible paths are structured in phases (or modes), where smooth differentiable constraints describe the feasible motions in each mode and feasible transitions (switches) between modes, akin to multi-modal path planning \cite{Hauser2011}.
A Logic-Geometric Program is a nonlinear mathematical program (NLP) of the form
\begin{subequations}\label{eq:rearrangement}
\begin{align} 
	\min_{K, s_{1:K} \atop x:[0,t_K]\to \XX } ~& \int^{t_K}_0 c(\bar x(t))~ dt \\
    \text{s.t.}~~ & ~~~~~~~~x(0) = x_0,~~~~\, h_\text{goal}(x(t_K))\leq 0, \\
    & \forall_k: \forall_t\in[t_{k\1}, t_k]: h_\text{mode}(\bar x(t), s_k) \leq 0, \\
    & ~~~~~~~~~~~~~~~\, h_\text{switch}(\bar x(t_k), s_{k\1}, s_k) \leq 0, \label{eq:rearrangement:constraint}\\
    & \forall_k: s_k \in \text{succ}(s_{k\1}), ~ s_K \models \mathfrak{g}
\end{align}
\end{subequations}

where $\bar x(t)=(x(t), \dot x(t), \ddot x(t))$. The decision variables of this NLP are the number of phases $K$, the symbolic mode sequence $s_{1:K}$, and the continuous path $x(t)$.
The path is constrained to start at $x_0$ and end in a configuration $x(t_K)$ consistent with the goal constraints $h_\text{goal}$.
In each phase, the path needs to fulfill mode constraints $h_\text{mode}$ depending on the symbolic mode $s_k$, and each $\bar x(t_k)$ needs to fulfill mode-switch constraints $h_\text{switch}$ depending on the symbolic mode transition.
The times $t_k$ of mode transitions are assumed to be fixed and equal spaced.
Only certain mode transitions are possible logically, here notated via a successor function (usually given as a STRIPS-like task planning domain\cite{FIKES1971}).
Finally, $\mathfrak{g}$ denotes a logical goal.
All constraint functions $h$ are assumed smoothly differentiable, and are here written as inequalities -- but this is meant to also include equality constraints which an NLP solver handles differently.

\subsection{Problem Formulation}
The LGP \eqref{eq:rearrangement} is a very general formulation for TAMP. In this work, we consider a subclass of problems where the available discrete actions are restricted to \text{pick} and \text{place} of any object $o_m$ in the scene using a stable grasp and placement on the ground. In all problems, the goal $\mathfrak{g}$ is to put the goal object $o_{\textit{goal}} \in \mathcal{O}$ on the red goal surface $goal$. We define this problem as $\mathcal{P}(x_0, \mathfrak{g})$ for initial configuration $x_0 \in \XX$. 

We consider sequential manipulation problems requiring multiple regrasps, moving away objects and moving objects through tight spaces. However, $\mathfrak{g}$ does not contain any information to plan these manipulations, grasps and movements.
Our approach is to find and solve auxiliary subproblems, which will guide our planner over a sequence of subgoals to a configuration $x_\text{feas}$ from which $\mathfrak{g}$ can be solved. %

\subsection{Pick-and-Place Auxiliary Subproblems}
Starting from a configuration $x_i \in \XX$, the \textit{pick-and-place auxiliary subproblem} $\mathcal{\tilde{P}}(x_i, g)$ is defined as the task of moving an object $o \in \mathcal{O}$ to a position $z \in SE(3)$ in a single pick-and-place sequence, with $g$ being the object-position tuple $(o, z)$ and defined as a \textit{subgoal}.

$\mathcal{\tilde{P}}$ is an instance of an LGP problem with two phases $K=2$ and a fixed sequence of mode-transitions: $(\texttt{pick} ~ o_1)$, $(\texttt{place} ~ o_1 ~ \texttt{in} ~ z)$. %
To pick an object, the robot should be able to touch it, with grasps possible from all angles. Once picked, the object is attached with a rigid joint until it is placed in $z$. 
The poses for pick and place are optimized jointly for the following constraints.
The goal constraint $h_\text{goal}$ is a \textit{pose constraint} depending on $z$. The switch constraints $h_\text{switch}$ are \textit{touch} and \textit{stable} for \texttt{pick}, and \textit{above} and \textit{stable-on} for the \texttt{place} action as defined in \cite{Toussaint2018}.

The solution of a subproblem $\mathcal{\tilde{P}}(x_i,g)$ is a collision-free path $X$ from $x_i=X(0)$ to a final configuration $x^f=X(T)$ that meets the subgoal $g$.
We use the following notation to indicate feasibility of a subproblem and return the final configuration:
\begin{equation}
x^f = \text{solve}~ \mathcal{\tilde{P}}(x_i, g),
\end{equation}

The configuration $x^f$ is a valid configuration that can be used later as a start for the original problem $\mathcal{P}(x^f, \mathfrak{g})$ or subsequent subproblems.
\subsection{Subproblem Sequence}
\begin{figure}
    \centering
    \includegraphics[width=.85\linewidth]{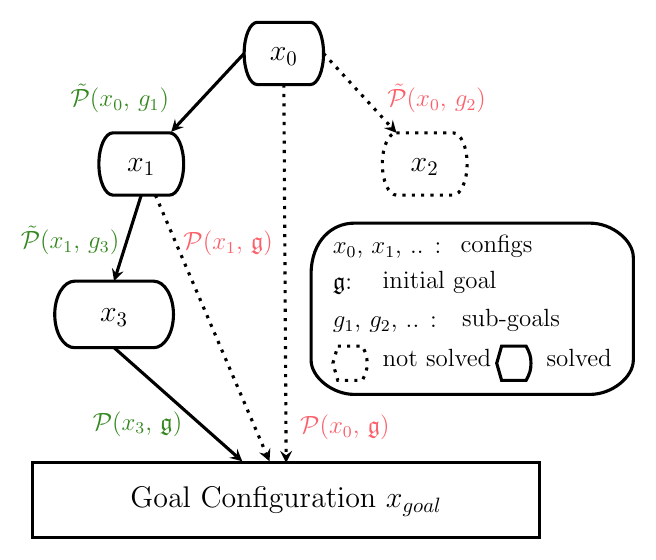}
    \caption{Subproblem sequence. Solved subproblems and problems in green, unsolved in red.}
    \label{fig:subproblems}
\end{figure}

To solve the original problem \eqref{eq:rearrangement}, we aim to find a sequence of subgoals $[g_1, \ldots,  g_L]$ such that,
\begin{align}
& x^f_0 \equiv x_0, \\
&  \forall_{i \in [1,...,L]}: x^f_i = \text{solve}~\mathcal{\tilde{P}}(x^f_{i-1}, g_{i}), \\ 
& ~~~~~~~~~~~~ x_\text{goal} = \text{solve}~ \mathcal{P}(x^f_L, \mathfrak{g}) ~. \end{align}
The sequence of subproblems is represented as a graph in Fig.\,\ref{fig:subproblems}, which starts at node $x_0$, and is expanded using new subproblems (e.g. $\mathcal{\tilde{P}}(x_0, g_0)$ and $\mathcal{\tilde{P}}(x_0, g_1)$) or the original problem $\mathcal{P}(x_0, \mathfrak{g}$). The solid edges represent solution paths, leading to new valid configurations. Once $\mathcal{P}$ is solved for $\mathfrak{g}$, we can trace the complete solution from $x_0$ to $x_\text{goal}$.

\section{Solving Multi-Step Manipulation Problems using a Sequence of Subgoals}
The main idea of our method is to perform a \textit{forward search over possible pick-and-place subproblems $\mathcal{\tilde{P}}$} to reach a configuration $x_f$, for which we can solve the original problem $\mathcal{P}(x_f, \mathfrak{g})$. %
We cannot derive such subproblems directly from the goal, because the symbolic goal only defines constraints on one object ($o_{goal}$) and these constraints are only related to its final position.
Most importantly, the goal does not provide information on whether we need to move other objects, in what order or over which trajectories.
Instead, we introduce heuristics in our search algorithm to come up with helpful and solvable subproblems.

In Alg.\,\ref{alg:search} we outline the main steps of our approach.
At every iteration of the main loop, we select one of the configurations $x$ from the list $L$ using feasibility heuristics in \texttt{select\_node}. We try to solve the original problem $\mathcal{P}(x, \mathfrak{g})$ with \texttt{solve},
which uses Multi-Bound Tree Search (MBTS)\cite{17-toussaint-ICRA}, which is a solver for LGP. It explores possible sequences of discrete actions using a tree search algorithm with multiple geometric bounds. These bounds allow to quickly test the feasibility of path and mode-switch configurations in \eqref{eq:rearrangement} for a fixed action skeleton by solving informative relaxations of the trajectory optimization problem. If successful, \texttt{solve} returns a path $X$.

If $\mathcal{P}(x, \mathfrak{g})$ was not solved, we try to reach configurations from which it may be solvable. We generate subgoals for objects reachable to the agent using \texttt{propose\_subgoals}, giving us a set of auxiliary pick-and-place subproblems $\mathcal{\tilde{P}}$.
In comparison to the original problem $\mathcal{P}$, solving $\mathcal{\tilde{P}}$ does not require a search over different discrete actions. We solve the auxiliary subproblems using a fixed-action-sequence solver \texttt{sub\_solve}: a combination of nonlinear optimization and sample based motion planning. First, we compute the mode-switch configurations for the grasp and the placement of the object using nonlinear optimization with a randomized initialization. Second, we use bi-directional Rapidly-exploring Random Trees (RRTs) \cite{Kuffner2000}, to compute a collision-free path between the mode-switch configurations, similar to \cite{Hartmann_2023}.

The solutions of $\mathcal{\tilde{P}}$ give us trajectories leading to new valid configurations, which are added to the list \textit{L} of start configurations to try.
The search concludes, if either we solve $\mathcal{P}$ or we do not have any more configurations to expand.

\begin{algorithm}[h] %
  \caption{Forward subproblem search}
\label{alg:search}
\begin{algorithmic}[1] %
	\State \textbf{Input:} Start configuration $x_0$, logical goal $\mathfrak{g}$
   \State $L \gets \{x_0\}$ 
   \Comment{Initialize list of valid configurations}
   \While {$L$ not empty}
   \State $x \leftarrow \texttt{select\_node}(L)$ \Comment{s. \ref{sec:select}}
	\State $X \leftarrow \texttt{solve} ~ \mathcal{P}(x, \mathfrak{g})$ 
   \Comment {Try to reach goal}
   \If {$X$ \texttt{found}}
     \State  \textbf{Return} $X_{\text{full}}$ \Comment{Trace solution back to $x_0$}
  \EndIf
  \For{$o \in \mathcal{O}$}	
  \If {not \texttt{reachable}($x$, $o$)  }
  \State \textbf{Continue}
  \EndIf
  \State $Z \leftarrow \texttt{propose\_subgoals}(x, o)$ 
  \Comment{s. \ref{sec:gensubgoals}}
	  \For {$z \in  Z $}
   	\State $g \leftarrow (o, z)$  \Comment{New subgoal $g$}
	\State $X \leftarrow \texttt{sub\_solve} ~ \mathcal{\tilde{P}}(x, g)$  
	\If {$X$ \texttt{found} and not \texttt{rej}($L$, $x^f$)}
    \State $\texttt{Insert}(L,x^f)$  
    \Comment{add last element of $X$}
	    \EndIf
	  \EndFor
	\EndFor
  \EndWhile
\State \textbf{Return} $\texttt{None}$ \Comment{No solution found}
  \end{algorithmic}
\end{algorithm}
\subsection{Prioritizing configurations with \texttt{select\_node}}
\label{sec:select}
Intuitively, we would like to prioritize configurations from which the LGP-Solver is more likely to solve the problem $\mathcal{P}$. %
We introduce the  score function
$s(x) = 10 v_o + 5 v_g + v_\text{dist}$  to select configurations, where
\begin{itemize}
\item $v_{o}$ is 1, if there is a line-of-sight between $o_\text{goal}$ and the goal, and 0 otherwise.
\item $v_{g}$ is 1, if there is a line-of-sight between $o_\text{goal}$ and the agent, and 0 otherwise.
\item $v_\text{dist}$ measures proximity $d$ of $o_\text{goal}$ to the goal, and is set to 5 if $d < 0.2$, 2 if $d < 0.4$, and 0 otherwise.
\label{heur}
\end{itemize}
We only expand a configuration twice, if all other configurations in $L$ have been tried. If every configuration has been tried twice, we stop the search.
\subsection{Generating Subproblems with \texttt{propose\_subgoals}}
\label{sec:gensubgoals}
First, to decide on the object $o$, we try to find a collision-free path to the object center with bi-directional RRT to check if the object is reachable to the agent. 
Hard to reach objects can still be deemed unreachable, as we limit the RRT step budget.

Second, for a given object $o$ we generate a set of subgoal positions $Z$ with the \texttt{propose\_subgoals} method:
\begin{enumerate}
	\item Generate position candidates.
	\item Filter candidates with \texttt{Filter} (see \ref{sec:sgselect}).
	\item From the filtered positions sample a small subset ($Z$).
\end{enumerate}

We evaluated several methods to generate the initial subgoal positions:
\begin{itemize}
	\item \texttt{Rnd}: randomly generated
	\item \texttt{Btl}: using bottleneck analysis
	\item \texttt{Hum}: from human demonstrations
\end{itemize}
\texttt{Btl}: to find optimal subgoal positions we look for \textit{bottlenecks} in the scene, i.e. entrances or narrows passages you need to pass to reach other areas. Similar to \textit{diverse density} analysis in \cite{McGovern2001AutomaticDO}, we look for areas with a higher number of possible paths leading through them.
First, we rasterize the scene and construct a grid-like graph, with vertices connected only if there is a path between them for the object $o$.
Then we calculate the shortest paths from vertices in a small area around the current object to all other vertices in the graph. We use the number of paths passing through a vertex relative to the number of vertices as a measure of \textit{paths density} of this vertex. %
Compared to random sampling, the \texttt{Btl} method is far more likely to propose a position inside a narrow space than inside a larger area, which helps to guide the subgoal generation to the bottlenecks.
\subsubsection{Filtering subgoal positions with \texttt{Filter}}
\label{sec:sgselect}
\begin{figure}[t]
\centering
    \begin{subfigure}[t]{.189\linewidth}
        \includegraphics[width=\linewidth]{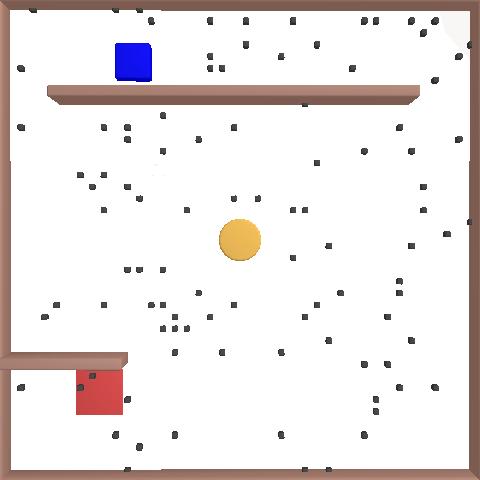}
        \caption{}
        \label{fig:subgoalsrndc}
    \end{subfigure}
	\hfill
    \begin{subfigure}[t]{.189\linewidth}
        \includegraphics[width=\linewidth]{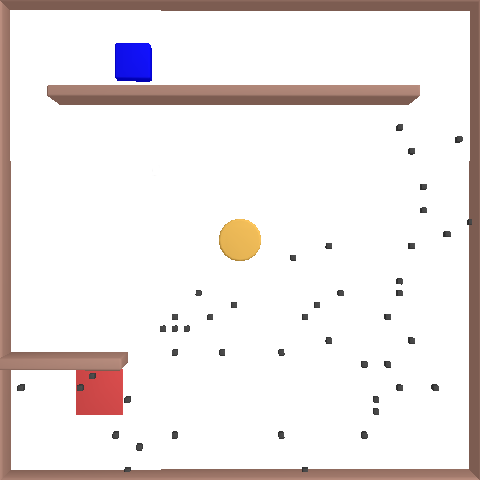}
        \caption{}
        \label{fig:subgoalsrndfc}
    \end{subfigure}
	\hfill
    \begin{subfigure}[t]{.189\linewidth}
        \includegraphics[width=\linewidth]{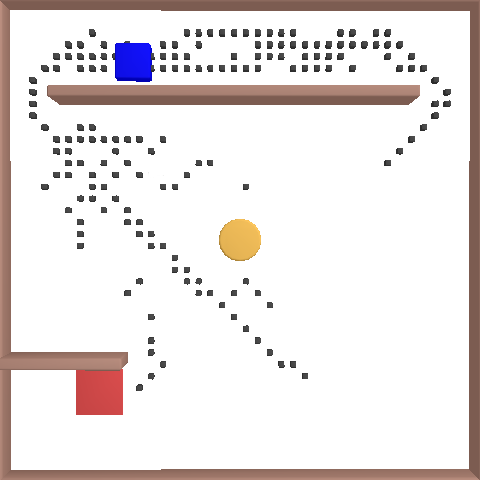}
        \caption{}
        \label{fig:subgoalspathc}
    \end{subfigure}
	\hfill
    \begin{subfigure}[t]{.189\linewidth}
        \includegraphics[width=\linewidth]{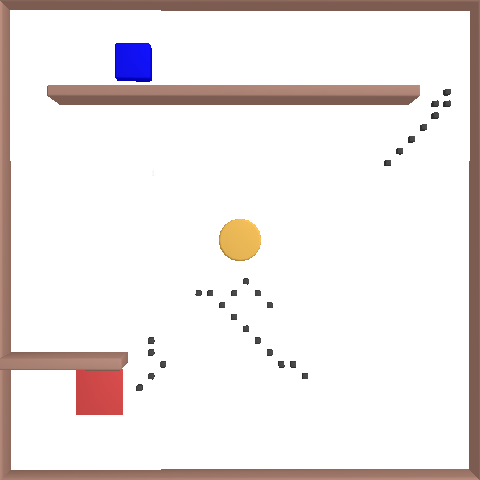}
        \caption{}
        \label{fig:subgoalspathfc}
    \end{subfigure}
	\hfill
    \begin{subfigure}[t]{.189\linewidth}
        \includegraphics[width=\linewidth]{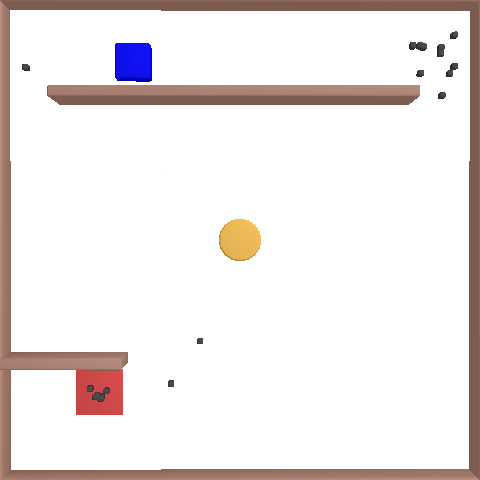}
        \caption{}
        \label{fig:subgoalshumc}
    \end{subfigure}
	\caption{Subgoals in the scene Wall. a) \texttt{Rnd} b) \texttt{Rnd} filtered by heuristics. c) \texttt{Btl} d) \texttt{Btl} after filtering e) \texttt{Hum} }
    \label{fig:subgoals2}
\end{figure}

from the set of generated positions, we select the most promising ones with heuristics from \ref{sec:select}: we calculate the feasibility score of a hypothetical solution configuration, assuming the object $o$ was successfully moved to the subgoal position $z$. Then we filter the positions based on their scores. 
Filtering helps to find positions, where $o_{goal}$ would be either close to the $goal$ or in line-of-sight of it.
A distribution of the subgoal positions before and after filtering is visualized in Fig.\,\ref{fig:subgoals2}. 

Finally, from the filtered positions we randomly sample a subset of subgoals, adding a trivial subgoal, where the position is equal to the current position of the object.%

\subsection{Rejecting similar configurations with \texttt{rej}}
After a subproblem is solved, we add the last configuration of the solution path to the list $L$.
To avoid repeating problems, we calculate a similarity value for configurations, based on the discretized positions of movable objects, and allow at most two similar configurations to be added.
\section{Experiments}
\begin{figure}[t]
  \subfloat[Corner]{\label{fig:i10}\includegraphics[width=0.22\linewidth]{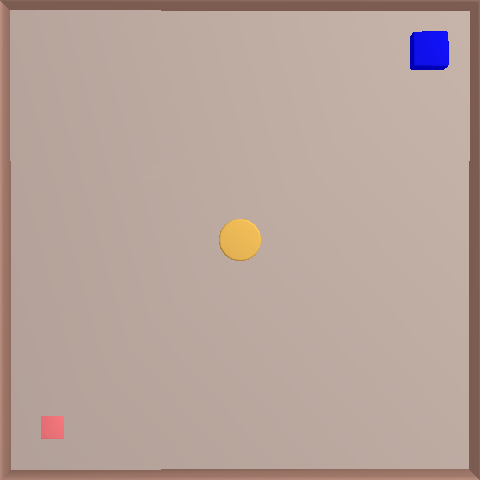}}\hfill
  \subfloat[2 blocks]{\label{fig:p1}\includegraphics[width=0.22\linewidth]{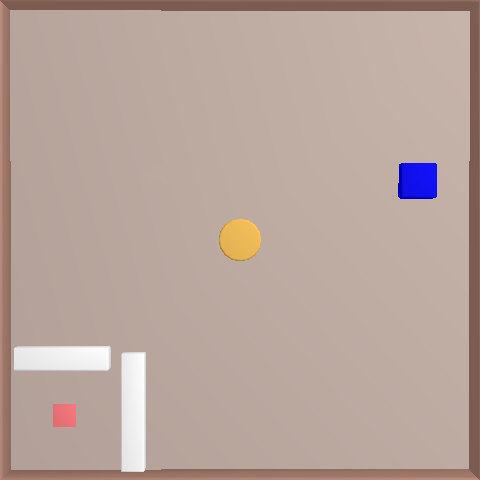}}\hfill
  \subfloat[4 blocks]{\label{fig:p6}\includegraphics[width=0.22\linewidth]{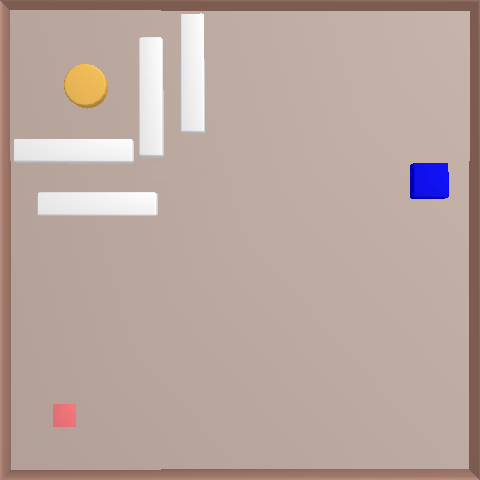}}\hfill
  \subfloat[O-Room]{\label{fig:i7}\includegraphics[width=0.22\linewidth]{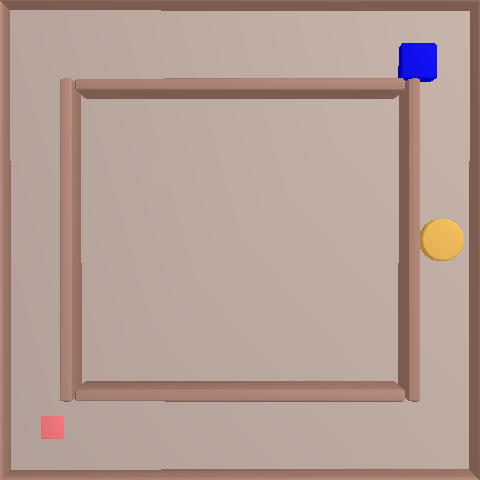}}\\
  \subfloat[Maze-Easy]{\label{fig:p2}\includegraphics[width=0.22\linewidth]{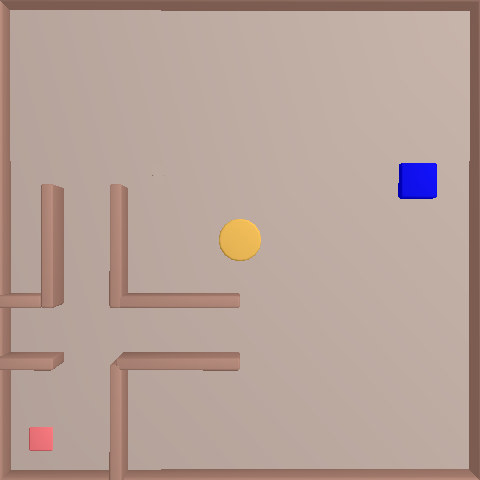}}\hfill
  \subfloat[Wall-Easy]{\label{fig:i5}\includegraphics[width=0.22\linewidth]{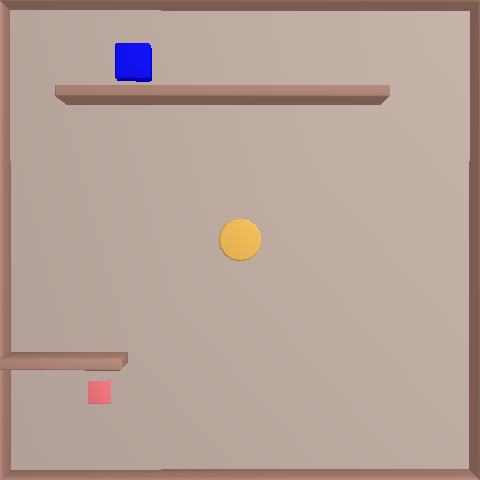}}\hfill
  \subfloat[Maze]{\label{fig:p3}\includegraphics[width=0.22\linewidth]{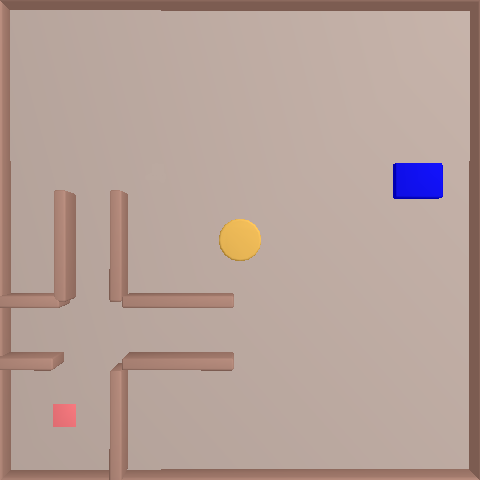}}\hfill
  \subfloat[Wall]{\label{fig:i6}\includegraphics[width=0.22\linewidth]{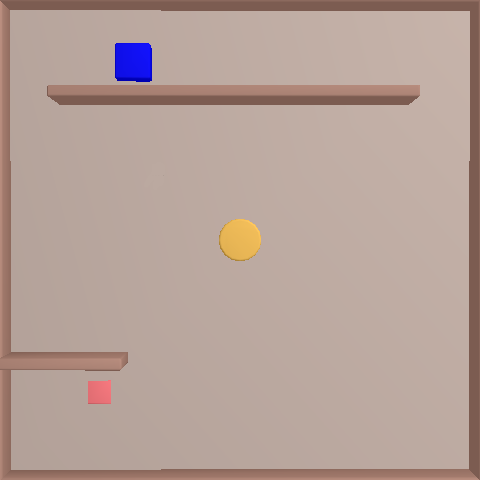}}%
\caption{Puzzle scenes: the agent (yellow circle) needs to bring the goal-object (blue) to the goal (red square). Movable obstacles in white, static walls in brown.}
\label{fig:problems3}
\end{figure}

\begin{figure}[t]
  \subfloat[a11-6]{\label{fig:p-sok0}\includegraphics[width=0.18\linewidth]{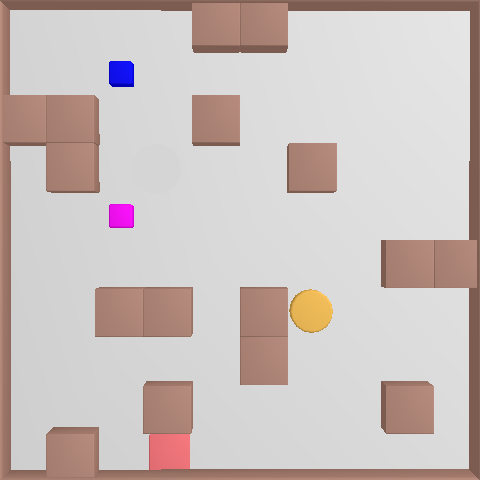}}\hfill
  \subfloat[a11-46]{\label{fig:p-sok1}\includegraphics[width=0.18\linewidth]{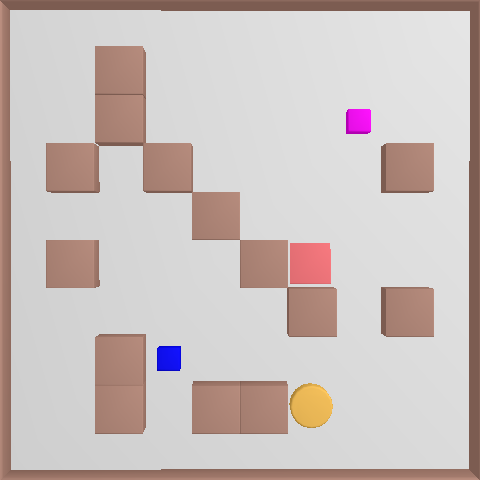}}\hfill
  \subfloat[a15-20]{\label{fig:p-sok2}\includegraphics[width=0.18\linewidth]{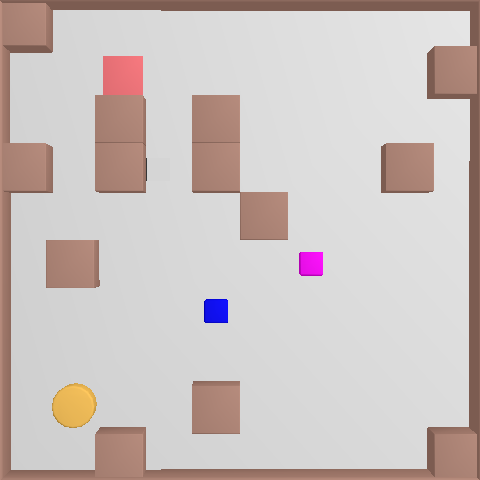}}\hfill
  \subfloat[a19-20]{\label{fig:p-sok3}\includegraphics[width=0.18\linewidth]{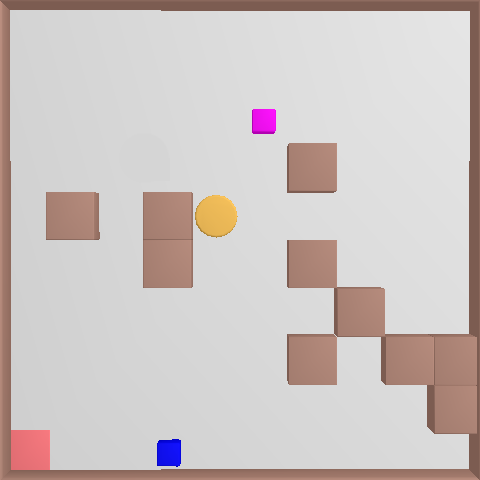}}\hfill
  \subfloat[a22-4]{\label{fig:p-sok4}\includegraphics[width=0.18\linewidth]{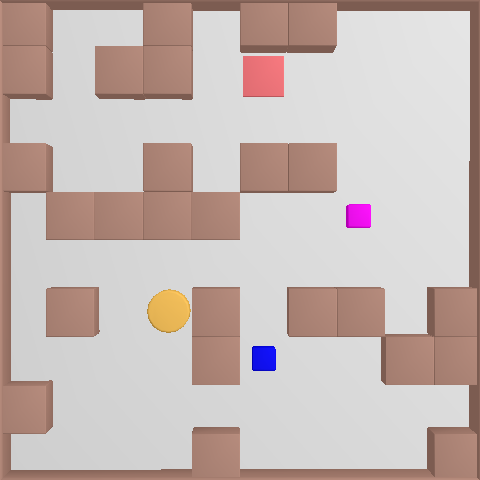}}\\
  \subfloat[a23-78]{\label{fig:p-sok5}\includegraphics[width=0.18\linewidth]{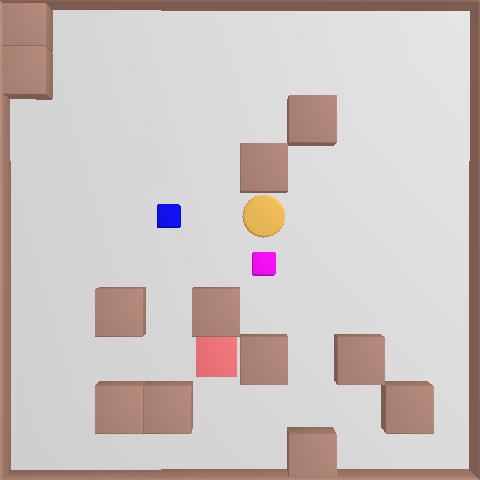}}\hfill
  \subfloat[a29-0]{\label{fig:p-sok6}\includegraphics[width=0.18\linewidth]{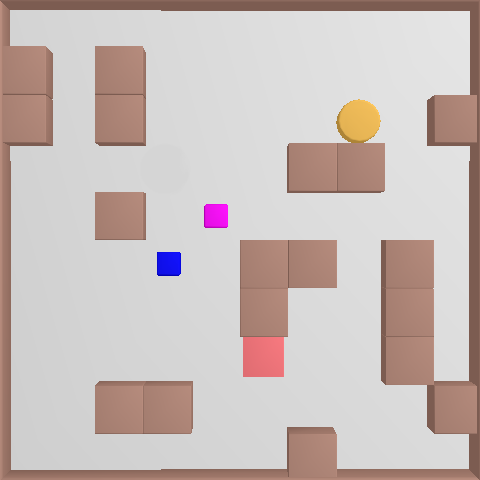}}\hfill
  \subfloat[a35-2]{\label{fig:p-sok7}\includegraphics[width=0.18\linewidth]{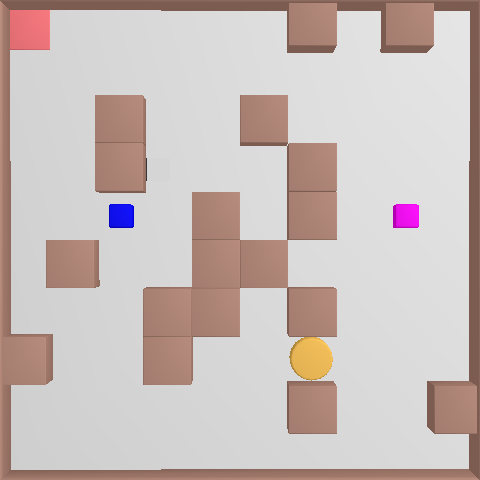}}\hfill
  \subfloat[a3-92]{\label{fig:p-sok8}\includegraphics[width=0.18\linewidth]{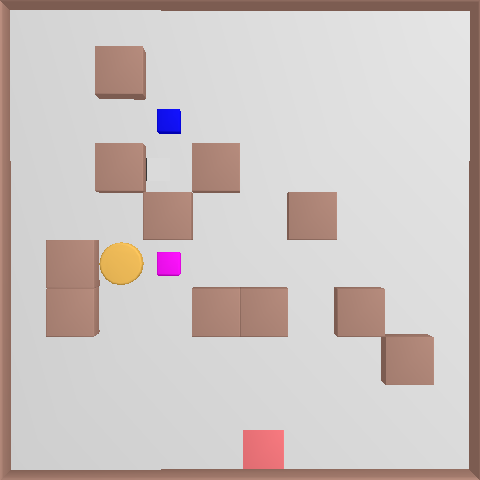}}\hfill
  \subfloat[a5-43]{\label{fig:p-sok9}\includegraphics[width=0.18\linewidth]{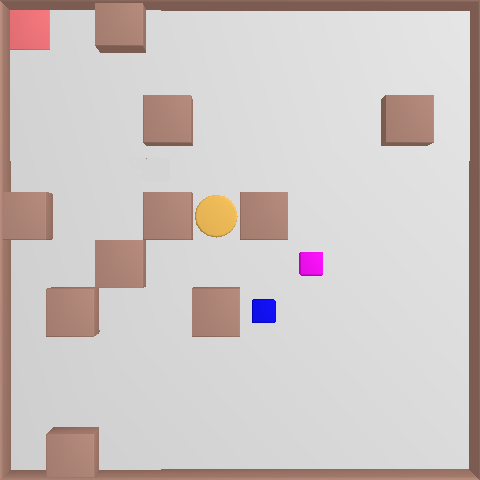}}\\
  \subfloat[a4-4]{\label{fig:p-sok10}\includegraphics[width=0.18\linewidth]{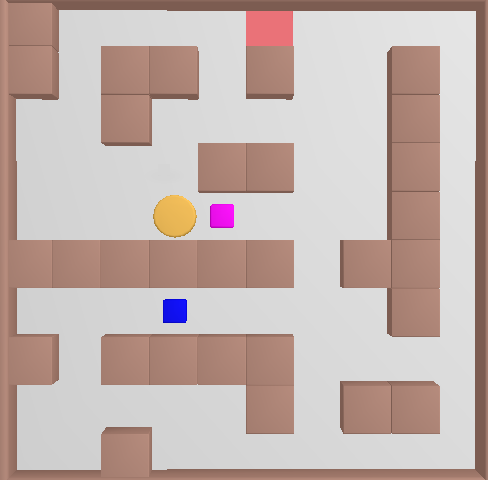}}\hfill
  \subfloat[a19-18]{\label{fig:p-sok11}\includegraphics[width=0.18\linewidth]{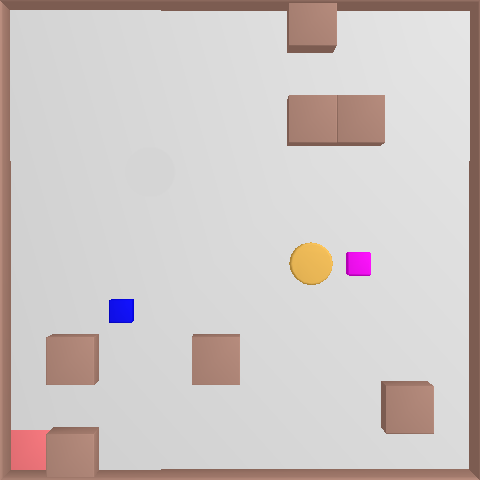}}\hfill
  \subfloat[a23-6]{\label{fig:p-sok12}\includegraphics[width=0.18\linewidth]{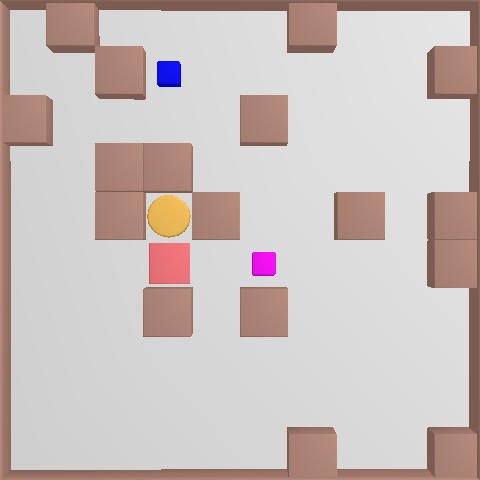}}\hfill
  \subfloat[a31-4]{\label{fig:p-sok13}\includegraphics[width=0.18\linewidth]{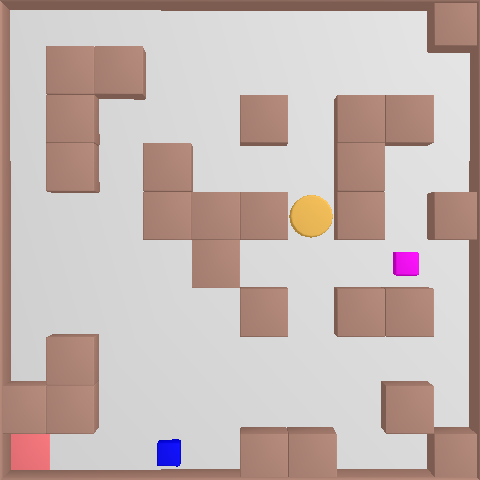}}\hfill
  \subfloat[a4-2]{\label{fig:p-sok14}\includegraphics[width=0.18\linewidth]{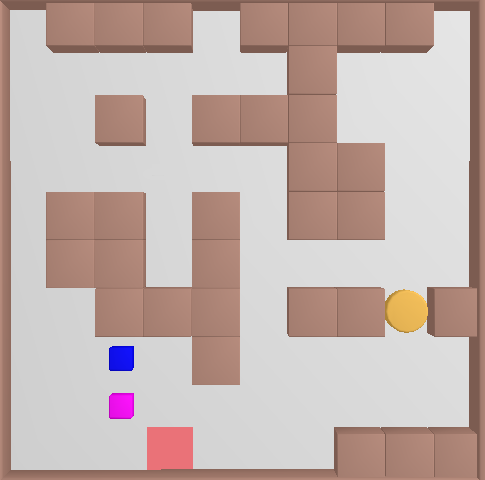}}%
\caption{Proceduraly Generated Scenes (subset) with walls (brown) and two movable objects: the goal-object (blue) and an obstacle-object (pink).}
\label{fig:problems}
\end{figure}

We evaluated the solver on pick-and-place puzzles of different complexity, addressing common TAMP challenges.
While we formulated our problem in a general way with object poses in $SE(3)$, we consider a robot with 2-DoF to highlight the challenges of interdependencies between grasps and trajectories.
Our benchmark problems include:\footnote{Available at https://github.com/svetlanalevit/manipulation-puzzles.} %
\begin{itemize}
    \item 9 manually designed puzzles, challenging for the solver due to multiple obstacles and bottlenecks, Fig.\,\ref{fig:problems3}.
\item 360 PCG (Procedural Content Generation) scenes: automatically generated maze-style puzzles with two objects, allowing us to evaluate how well the approach performs in general settings, Fig.\,\ref{fig:problems}. 
\end{itemize}

To create the PCG scenes, we used the PCGRL Framework, which can generate solvable gridworld game levels of varying difficulty\cite{Stolz2023}\cite{khalifa2020pcgrl}.
We converted the gridworld scenes (Sokoban levels) to continuous 3D TAMP problems. %

We measured the following metrics: 
\begin{enumerate}
\item \textit{Completeness}: solution ratio over multiple runs.
\item \textit{Performance}: solution time. %
\item \textit{Success rate}: percentage of solved PCG scenes.
\end{enumerate}

The \textit{performance} and \textit{completeness} experiments are repeated 10 times on 9 manually designed puzzles and 12 PCG problems. In Table\,\ref{tab:feas} we report the solution times, the standard deviation and the solution rate, if it is below 100\%. We used a single i7-11700@2.5\,GHz CPU for the tests. 
The \textit{success rate} experiments are repeated once on all PCG problems and we count the number of solved problems.

We compared three methods to generate subgoals (s. \ref{sec:gensubgoals}): \texttt{Rnd}, \texttt{Btl}, and \texttt{Hum}. Using random subgoals \texttt{Rnd}, we evaluated the importance of the individual components: 
\begin{itemize}
    \item \textbf{F} - \texttt{Filter}, heuristics based filtering of subgoals during subgoal generation, s. \texttt{Filter} in \ref{sec:sgselect}. 
    \item \textbf{P} - \texttt{Prio}: heuristics based selection of next configurations for node expansion in \texttt{select\_node}. 
    \item \textbf{R} - \texttt{Reject}: diversity based rejection of new configurations from solved subproblems.
\end{itemize}
In total, we evaluated the following combinations of methods (as reported in Tab.\,\ref{tab:feas}):
\begin{enumerate}
\item \texttt{Rnd+FPR}: the full method with random subgoals (\texttt{Filter}: yes, \texttt{Prio}: yes, \texttt{Reject}: yes).%
\item \texttt{Rnd+FP}: without rejection of similar configrations (\texttt{Filter}: yes, \texttt{Prio}: yes, \texttt{Reject}: \underline{no}).
\item \texttt{Rnd+FR}: without heuristics for configuration prioritization (\texttt{Filter}: yes, \texttt{Prio}:\underline{no}, \texttt{Reject}: yes)
\item \texttt{Rnd+R}: without heuristics (\texttt{Filter}: \underline{no}, \texttt{Prio}: \underline{no}, \texttt{Reject}: yes).
\item \texttt{Btl+FPR}: subgoals from the bottleneck analysis, (\texttt{Filter}: yes, \texttt{Prio}: yes, \texttt{Reject}: yes)
\item \texttt{Hum+R}: manual subgoals, without heuristics (\texttt{Filter}: \underline{no}, \texttt{Prio}: \underline{no}, \texttt{Reject}: yes)
\item \texttt{MBTS}: baseline method, Multi-Bound Tree Search.
\end{enumerate}

\section{Results}

\begin{table*}[ht] 
 \caption{Average solution time over 10 runs (s). \texttt{Rnd}: random subgoals, \texttt{Btl}: bottleneck subgoals, \texttt{Hum}: subgoals from human demonstrations. \ Ablations: \texttt{Rnd+FP}: no rejection of similar configs, \texttt{Rnd+FR}: no heuristics to select next config, \texttt{Rnd+R}: no heuristics.\,Solution rate in percent, if below 100\% and '-', if not solved. For \texttt{Btl+FPR}: in square brackets the time without computing subgoals.}
 \label{tab:feas} 
 \begin{center} \begin{tabular}{|c|c|c|c|c|c|c|c|}\hline 
 Scene&Rnd+FPR&Rnd+FP&Rnd+FR&Rnd+R&Btl+FPR&Hum+R&MBTS\\ 
\hline 
Cube-Free& 0.2{\tiny $\pm$0.01}& 0.2{\tiny $\pm$0.01}& \textbf{0.2}{\tiny $\pm$0.01}& 0.2{\tiny $\pm$0.01}& 0.2{\tiny $\pm$0.01} && 0.2{\tiny $\pm$0.01}\\ 
\hline 
Corner& 0.3{\tiny $\pm$0.01}& 0.3{\tiny $\pm$0.01}& \textbf{0.3}{\tiny $\pm$0.01}& 0.3{\tiny $\pm$0.01}& 0.3{\tiny $\pm$0.01} && 0.3{\tiny $\pm$0.01}\\ 
\hline 
2 blocks& 2.5{\tiny $\pm$0.01}& 2.6{\tiny $\pm$0.01}& 2.5{\tiny $\pm$0.01}& 2.5{\tiny $\pm$0.04}& \textbf{2.5}{\tiny $\pm$0.01} && 2.5{\tiny $\pm$0.02}\\ 
\hline 
4 blocks& 38.4{\tiny $\pm$23.01}& 35.2{\tiny $\pm$20.57}& 41.7{\tiny $\pm$15.45}& \textbf{30.4}{\tiny $\pm$23.22}& 165.8{\tiny $\pm$81.87}{\tiny [55.8]}& 176.6{\tiny $\pm$68.29}& -\\ 
\hline 
Maze-Easy& \textbf{54.0}{\tiny $\pm$10.96}& 56.0{\tiny $\pm$10.91}& 84.7{\tiny $\pm$38.45}& 155.4{\tiny $\pm$86.73}& 143.9{\tiny $\pm$94.29}{\tiny [85.7]}{\tiny (80.0\%)}& 55.2{\tiny $\pm$9.27}& -\\ 
\hline 
Wall-Easy& \textbf{1.8}{\tiny $\pm$0.16}& 2.1{\tiny $\pm$0.29}& 7.8{\tiny $\pm$9.89}& 9.4{\tiny $\pm$8.73}& 8.3{\tiny $\pm$2.19}{\tiny [2.4]}& 3.5{\tiny $\pm$4.45}& -\\ 
\hline 
O-Room& 68.3{\tiny $\pm$56.10}& 100.7{\tiny $\pm$80.62}& 92.9{\tiny $\pm$76.65}& 254.5{\tiny $\pm$170.08}& 259.2{\tiny $\pm$128.40}{\tiny [102.6]}& \textbf{61.1}{\tiny $\pm$30.37}& -\\ 
\hline 
Maze& 926.2{\tiny $\pm$615.78}& 1750.7{\tiny $\pm$1075.61}{\tiny (90.0\%)}& 1454.8{\tiny $\pm$1133.81}& 2867.7{\tiny $\pm$1781.67}{\tiny (40.0\%)}& 1772.5{\tiny $\pm$1319.81}{\tiny [839.4]}{\tiny (80.0\%)}& \textbf{154.9}{\tiny $\pm$162.50}{\tiny (80.0\%)}& -\\ 
\hline 
Wall& 190.1{\tiny $\pm$130.60}& 125.6{\tiny $\pm$111.11}& 240.0{\tiny $\pm$108.18}& 91.8{\tiny $\pm$84.86}& 59.7{\tiny $\pm$25.21}{\tiny [32.6]}{\tiny (90.0\%)}& \textbf{55.9}{\tiny $\pm$41.72}& -\\ 
\hline 
a11-6& 33.4{\tiny $\pm$9.71}& \textbf{31.6}{\tiny $\pm$4.61}& 56.3{\tiny $\pm$35.00}& 123.8{\tiny $\pm$94.97}& 67.7{\tiny $\pm$21.98}{\tiny [33.1]} && -\\ 
\hline 
a15-20& \textbf{11.2}{\tiny $\pm$1.08}& 12.0{\tiny $\pm$3.19}& 18.7{\tiny $\pm$9.77}& 57.2{\tiny $\pm$44.36}& 42.0{\tiny $\pm$15.49}{\tiny [14.2]} && -\\ 
\hline 
a19-20& 0.4{\tiny $\pm$0.01}& 0.4{\tiny $\pm$0.01}& \textbf{0.3}{\tiny $\pm$0.01}& 0.3{\tiny $\pm$0.01}& 0.3{\tiny $\pm$0.01} && 0.4{\tiny $\pm$0.01}\\ 
\hline 
a22-4& \textbf{45.9}{\tiny $\pm$0.22}{\tiny (20.0\%)}& 46.7{\tiny $\pm$0.32}{\tiny (20.0\%)}& 147.7{\tiny $\pm$91.98}& 120.9{\tiny $\pm$66.96}& 170.5{\tiny $\pm$54.49}{\tiny [81.4]}{\tiny (30.0\%)}& 144.4{\tiny $\pm$72.16}& -\\ 
\hline 
a23-78& \textbf{13.9}{\tiny $\pm$5.03}& 22.5{\tiny $\pm$27.59}& 48.5{\tiny $\pm$56.25}& 135.1{\tiny $\pm$100.35}& 60.2{\tiny $\pm$28.71}{\tiny [22.4]} && -\\ 
\hline 
a29-0& 31.9{\tiny $\pm$18.89}& 29.2{\tiny $\pm$5.11}& \textbf{28.1}{\tiny $\pm$11.44}& 47.1{\tiny $\pm$34.09}& 46.7{\tiny $\pm$1.95}{\tiny [22.5]} && -\\ 
\hline 
a35-2& \textbf{9.8}{\tiny $\pm$2.32}& 10.6{\tiny $\pm$1.75}& 14.3{\tiny $\pm$8.99}& 60.9{\tiny $\pm$55.65}& 55.5{\tiny $\pm$24.94}{\tiny [17.6]} && -\\ 
\hline 
a3-92& 30.2{\tiny $\pm$21.01}& \textbf{29.3}{\tiny $\pm$32.81}& 46.3{\tiny $\pm$23.11}& 99.4{\tiny $\pm$70.47}& 48.2{\tiny $\pm$22.97}{\tiny [20.0]}& 62.9{\tiny $\pm$76.90}& -\\ 
\hline 
a5-43& \textbf{21.1}{\tiny $\pm$5.47}& 29.3{\tiny $\pm$24.47}& 41.0{\tiny $\pm$16.94}& 67.3{\tiny $\pm$35.52}& 36.6{\tiny $\pm$7.27}{\tiny [16.5]} && -\\ 
\hline 
a19-18& 15.7{\tiny $\pm$4.15}& \textbf{14.2}{\tiny $\pm$5.41}& 21.4{\tiny $\pm$11.76}& 47.1{\tiny $\pm$28.99}& 42.1{\tiny $\pm$23.61}{\tiny [16.7]} && -\\ 
\hline 
a23-6& 73.1{\tiny $\pm$42.11}& \textbf{61.6}{\tiny $\pm$39.50}& 148.3{\tiny $\pm$102.77}& 86.9{\tiny $\pm$47.62}& 84.4{\tiny $\pm$35.56}{\tiny [37.4]}& 69.8{\tiny $\pm$60.94}& -\\ 
\hline 
a31-4& \textbf{22.0}{\tiny $\pm$3.86}& 24.0{\tiny $\pm$12.99}& 25.2{\tiny $\pm$12.94}& 43.1{\tiny $\pm$15.09}& 50.8{\tiny $\pm$3.41}{\tiny [20.5]} && -\\ 
 \hline 
 \end{tabular} 
 \end{center}
 \end{table*}
\textbf{Comparison to the MBTS Baseline}:
Table\,\ref{tab:feas} summarizes the results of our evaluations. We find that our method (Rnd+FPR) consistently solves the benchmark problems while the baseline solver (MBTS) fails (dashes) at problems with objects hidden behind walls and narrows passages (e.g. Maze, Maze-Easy, O-Room). As our method tries direct MBTS first, the runtimes for solved problems are the same.

We further analyzed the \textit{success rate} over all 360 PCG problems and found 99.7\% for Rnd+FPR vs 73.6\% for MBTS. The improvement is especially visible for problems where $o_\text{goal}$ is behind obstacles: in 122 out of 360 PCG scenes the $o_\text{goal}$ is at least partially occluded from the agent by an obstacle, while the other 238 have no occlusion (direct line-of-sight). With the subgoals search, we can solve 100\% of non-occluded and 99.1\% of occluded problems. With MBTS 88.24\% and 45.08\% respectively.  %
The scene a4-2 (Fig.\,\ref{fig:p-sok14}) was not solved, as all objects were marked as unreachable. %

\begin{figure}[t]
\centering
    \begin{subfigure}[t]{.49\linewidth}
        \includegraphics[width=\linewidth]{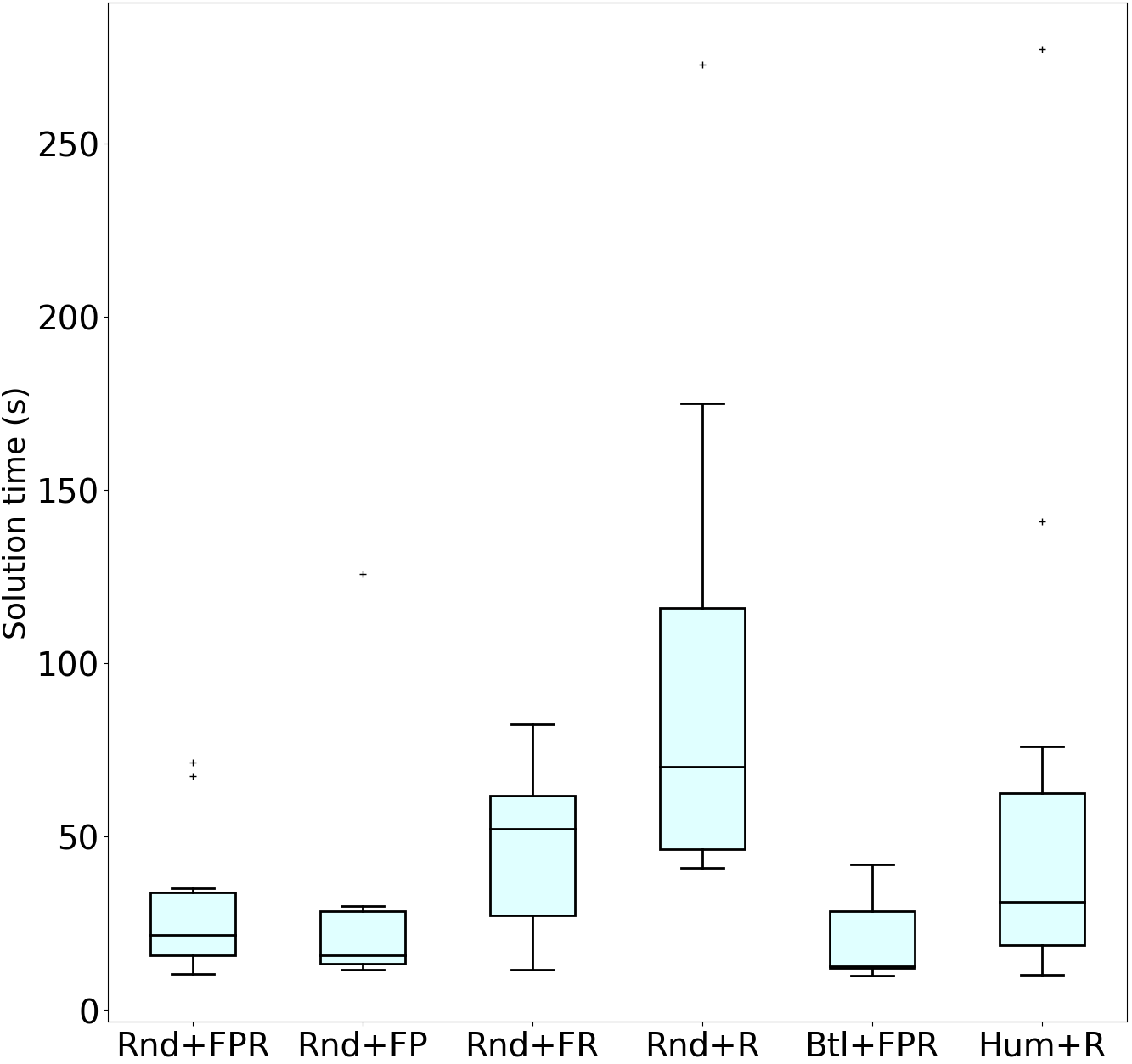}
        \caption{PCG Scene a3-92}
        \label{fig:boxplots-3-92}
    \end{subfigure}
        \hfill
    \begin{subfigure}[t]{.49\linewidth}
        \includegraphics[width=\linewidth]{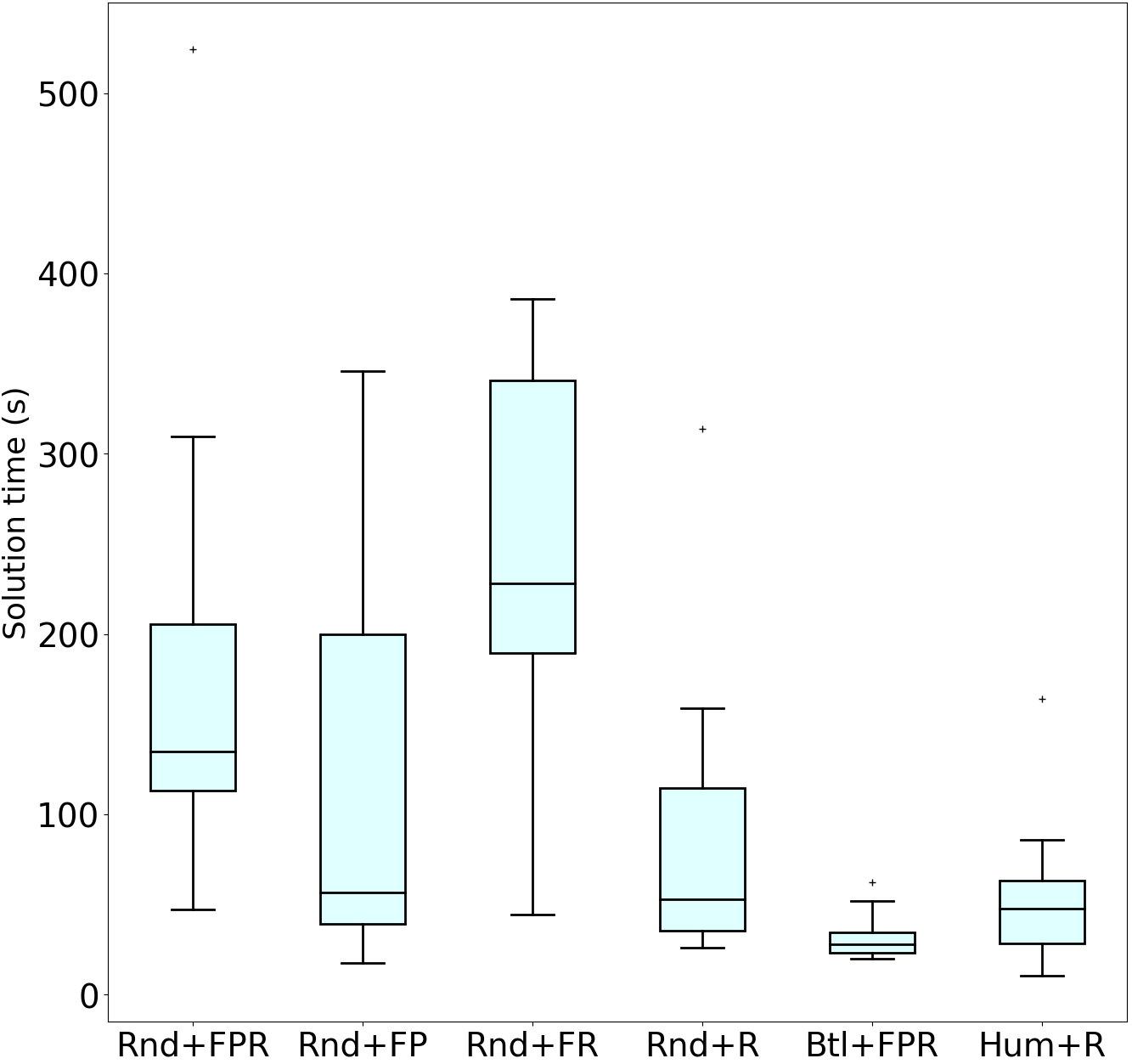}
        \caption{Wall}
        \label{fig:boxplots-wall}
    \end{subfigure}
        \hfill
	\caption{Runtime comparison over 10 runs (excl. runs without a solution). For \texttt{Btl}: the time without subgoal generation.  }
    \label{fig:boxplots}
\end{figure}

\textbf{Comparison of the subgoal generation methods}:
Table\,\ref{tab:feas} also reports the results for the three subgoal generation methods described in \ref{sec:gensubgoals} in the columns \texttt{Rnd+FPR}, \texttt{Btl+FPR} and \texttt{Hum+R}.
Overall, the \texttt{Rnd+FPR} (random subgoals with heuristics) performs better than \texttt{Btl} subgoals in terms of solution time and solution rate for most scenes: it shows faster average time in 16 out of 21 scenes, and similar in 4, and has in all but one scenes 100\% solution rate. The observed time difference is mostly due to the overhead to calculate the bottleneck subgoals. However, for scenes Wall and Maze, where the subgoals need to be placed in specific small areas, the \texttt{Btl} and \texttt{Hum} subgoals lead to faster search times (excluding the time to create or generate subgoals), as can be seen in Fig.\,\ref{fig:boxplots-wall}. 

\textbf{Discussion of feasibility heuristics}:
Filtering subgoals with \texttt{Filter} provides a significant performance boost, as reported in columns \texttt{Rnd+FR} and \texttt{Rnd+R}. While we sample the same amount of subgoals in both methods, with filtering we select subgoal positions with $o_{goal}$ closer to the goal or in line-of-sight of it.
However, as seen in Fig.\,\ref{fig:boxplots}, filtering of random subgoals leads to highly varying results in the Wall scene, where success strongly depends on finding the narrow opening, while filtering leaves only few candidates in this area (Fig.\,\ref{fig:subgoalsrndfc}).

Prioritizing configurations with a higher score, as described in Sec.\,\ref{sec:select}, leads to a faster solution in 16 out of 21 scenes, as can be seen on \texttt{Rnd+FPR} and \texttt{Rnd+FR} results of Tab.\,\ref{tab:feas} and both scenes in Fig.\,\ref{fig:boxplots}.

\textbf{Rejecting similar configurations}:
We compare the runtime of the full version \texttt{Rnd+FPR} to the ablation version \texttt{Rnd+FP}, where we do not reject similar configurations.
In most scenes the search finishes before we encounter similar configurations, except for Maze, Wall and O-Room, which take longer. For Maze we observed a higher number of rejected configurations in \texttt{Rnd+FPR}, and for this problem the non-rejecting ablation \texttt{Rnd+FP} is slower. However, for the scene Wall, we see the opposite effect in some of the runs, with the lower median time for \texttt{Rnd+FP} (Fig.\,\ref{fig:boxplots-wall}).
This indicates that it can be beneficial to try similar configurations for problems, where the objects are initially hard to reach.

\textbf{Comparison to human-baseline subgoals}:
Only two scenes (Maze and Wall) significantly benefit from the \texttt{Hum} subgoals, and the scene O-Room is slightly better than \texttt{Rnd+FPR}. These are hard problems with narrow passages, but only a single object without additional obstacles. We conclude that having a \textit{good initial guess} for the subgoals only helps if the scene remains unchanged and has only few possible solutions. In most other problems (especially 4-blocks and the PCG scenes) the positions of the agent and the objects can be very different in every solution. Therefore, \textit{generating subgoals dynamically for every configuration leads to better average performance than having a set of good but fixed subgoals}.

\section{Conclusions}
Integrating subproblem search with a baseline LGP-Solver enables us to plan for scenes which are not feasible for the baseline solver alone: containing occluded objects, obstacles and multiple narrow passages. By using simple heuristics to generate subgoals and select configurations, we can significantly decrease the time to solution.
We have shown that our approach is applicable to a wide variety of layouts.

For most scenes, the subgoals generated by the random heuristic lead to similar or better solution rates and times compared to subgoals from human demonstrations.
This shows that our approach does not require the subgoals to be in optimal positions, but it benefits from the ability to dynamically generate diverse subgoals.
The two scenes (Maze and Wall) where manual subgoals provide an advantage, are also the ones where a bottleneck heuristic performs better.

Our approach results in very interesting puzzle solutions, combining object repositioning, navigating through passages and refined multi-step object interactions.
However, the solutions could be potentially optimized in terms of the path lengths or the number of object interactions. 

In our evaluations we focused on pick-and-place actions, but the approach is also applicable to other interactions, as pushing.
\bibliographystyle{IEEEtran}
\balance
\bibliography{IEEEabrv,bib/general.bib}
\end{document}